\pdfoutput=1
\documentclass[journal]{IEEEtran}
\usepackage{xspace}
\usepackage{booktabs}
\usepackage{amsmath}
\usepackage{amssymb}
\usepackage{algorithmic}
\usepackage{algorithm}
\usepackage{multirow}
\usepackage{siunitx}
\usepackage{graphicx}
\usepackage{enumitem}
\usepackage{url}       %
\usepackage{newtxtext} %
\usepackage{tabularx} %

\usepackage{color_tudelft}
\usepackage{tikz}
\usetikzlibrary{arrows.meta,calc}
\definecolor{netc}{HTML}{D9732B}      %
\definecolor{corc}{HTML}{C0392B}      %
\colorlet{obsc}{corc}
\definecolor{datac}{HTML}{2A78D6}     %
\definecolor{flowgray}{HTML}{7A8699}  %
\usepackage{totcount}
\regtotcounter{todocount}
\newcounter{todocount}

\newcommand{\todowarning}{%
  \ifnum\totvalue{todocount}>0
    \textcolor{red}{\textbf{Warning: \total{todocount} TODO(s) remaining!}}%
  \else
    \textcolor{green}{\textbf{All TODOs completed!}}%
  \fi
}

\definecolor{todoblue}{HTML}{4A90E2}
\definecolor{todored}{HTML}{E74C3C}
\definecolor{todogreen}{HTML}{27AE60}
\definecolor{todoyellow}{HTML}{F39C12}
\definecolor{todopurple}{HTML}{8E44AD}

\newcolumntype{Y}{>{\centering\arraybackslash}X}

\newcommand{\energydiff}{\textit{EnergyDiff}\xspace}
\newcommand{\ours}{\textit{SmartMeterFM}\xspace}
\newcommand{\sssd}{SSSD$^{\text{S4}}$\xspace}

\DeclareMathOperator{\ple}{PLE}
\DeclareMathOperator{\layernorm}{LayerNorm}

\newcommand{\Dd}{\mathrm{d}}

\title{SmartMeterFM: Flow Matching for Smart Meter Data Generation, Imputation, and Super-Resolution Without Retraining}
\author{Nan Lin,~\IEEEmembership{Graduate Student Member,~IEEE}, Yanbo Wang, Jacco~Heres,\\ Peter~Palensky,~\IEEEmembership{Senior Member,~IEEE}, Pedro~P.~Vergara,~\IEEEmembership{Senior Member,~IEEE}
\thanks{Nan Lin is funded by NWO Align4Energy Project NWA.1389.20.251.}%
\thanks{Nan Lin, Peter Palensky, and Pedro P. Vergara are with the Intelligent Electrical Power Grids Group, Delft University of Technology, 2628 CD Delft, The Netherlands; Yanbo Wang is with the Signal Processing Systems Group, Delft University of Technology; Jacco Heres is with Alliander N.V., 6812 AH Arnhem, The Netherlands (e-mail: N.Lin@tudelft.nl, P.Palensky@tudelft.nl, P.P.VergaraBarrios@tudelft.nl, Y.Wang-27@tudelft.nl, jacco.heres@alliander.com).}%
\thanks{The code for models and experiments used in this paper can be found in Github Repositories: https://github.com/sentient-codebot/SmartMeterFM and https://github.com/distributionnetworksTUDelft/SmartMeterFM}%
}

\newtheorem{proposition}{Proposition}

\begin{document}
\maketitle

\begin{abstract}
Distribution system operators need realistic load and generation profiles when
measurements are unavailable, incomplete, or too coarse for the intended analysis.
Training separate models for synthetic data generation, imputation, and
super-resolution can duplicate modeling effort as observation requirements change.
This paper presents \ours, a conditional flow matching model trained once on
monthly smart meter profiles at \SI{15}{\min} resolution. Customer and calendar
attributes condition the model during training, while numerical profile constraints,
observed intervals, and coarse measurements are incorporated through
projection-based guidance at inference. The same weights serve all tasks without task-specific retraining and
preserve temporal patterns and extreme values while producing profiles
consistent with the supplied observations.
A large-scale case study using Dutch consumption and photovoltaic generation
profiles reports lower mean Continuous Ranked Probability Score (CRPS) than the
selected task-specific baselines for imputation and 16-fold super-resolution.
On a public dataset of 2,400 residential meters, the same weights match the imputation CRPS of a task-trained diffusion model when sampled ancestrally, attain a 4.5\% lower super-resolution CRPS with the default sampler, and meet the specified monthly maximum and mean power within \SI{0.01}{\kilo\watt}.
\end{abstract}

\section{Introduction}
\label{sec:introduction}
Distribution system operators (DSOs) use load and generation profiles to assess
network loading and characterize the variability of electricity demand and
distributed generation~\cite{wang2019ReviewSmartMeter}. The usefulness of these
profiles depends on their temporal coverage and resolution. Missing intervals
can conceal peaks, and coarse measurements can obscure fluctuations within an
aggregation window. Privacy and commercial sensitivity also limit access to
measured customer profiles~\cite{asghar2017SmartMeterData}. These conditions
create a need for realistic synthetic profiles, estimates of missing values,
and finer-resolution profiles that respect the measurements available to the
operator.

The observations available to a DSO vary with the metering arrangement and
define the tasks. For a customer without a smart meter, monthly billing
records may provide total energy consumption in kWh but no \SI{15}{\min}
power readings. In \textit{conditional generation}, a model generates a
profile for specified customer and calendar attributes;
\textit{constrained generation} additionally matches this energy total and
recorded monthly power peaks in kW.
When parts of a smart meter record are missing, \textit{imputation} estimates
the missing intervals while retaining the measured values. When only coarse
interval readings are available, \textit{super-resolution} generates finer-resolution
profiles whose aggregates agree with those readings.
The available information generally admits multiple compatible profiles.
A probabilistic model represents this remaining variability, making profile
fidelity and consistency with supplied observations common requirements
across all tasks.

Meeting these requirements for monthly smart meter data demands modeling
temporal structure over several scales. At \SI{15}{\min} resolution, a month
contains approximately $3000$ time steps, a sequence length that makes
generative modeling of energy time series
demanding~\cite{lin2025EnergyDiffUniversalTimeseriesa}. These profiles combine
daily and weekly patterns, changes in activity, and occasional
extremes~\cite{wang2019ReviewSmartMeter}. Month lengths vary, and the
first day of a month shifts relative to the weekly cycle. High-fidelity
monthly modeling must account for these calendar effects while preserving
both short-duration fluctuations and the structure of the full profile.

The underlying monthly profile structure is common to these tasks.
Training separate models for each task or observation configuration, as in
dedicated imputation~\cite{tashiro2021CSDIConditionalScorebased,li2024LoadProfileInpainting}
and super-resolution models~\cite{song2022ProfileSRGANGANBased}, repeats
the effort of learning that structure. This motivates a shared model that
learns smart meter patterns once and adapts to the information available
in each setting.

Existing work offers complementary approaches to generation and reconstruction.
Bottom-up models simulate demand from occupancy and appliance
use~\cite{richardson2010DomesticElectricityUse,mckenna2016HighresolutionStochasticIntegrated},
and conditional copulas model dependencies in smart meter
profiles~\cite{duque2021ConditionalMultivariateElliptical}.
Neural generative approaches include conditional variational autoencoders
for multivariate load states~\cite{wang2022GeneratingMultivariateLoad},
generative adversarial networks for renewable
scenarios~\cite{chen2018ModelfreeRenewableScenario}, and diffusion models for
energy time series, customer-specific load profiles, and synthetic net
load~\cite{lin2025EnergyDiffUniversalTimeseriesa,wang2023CustomizedLoadProfiles,zhang2025GeneratingSyntheticNet}.
Such models have been compared for transformer- and consumer-level load
profiles~\cite{xia2024ComparativeAssessmentGenerative}.
For reconstruction, LoadPIN restores missing load segments and estimates
baselines~\cite{li2024LoadProfileInpainting}, while ProfileSR-GAN generates
high-resolution load profiles from coarse
measurements~\cite{song2022ProfileSRGANGANBased}.
Each of these models addresses generation or a single reconstruction task.
The DSO settings introduced above instead require realistic profile generation
and reconstruction from changing observations in one model.

A natural way to combine these capabilities is to separate the learned profile
distribution from the observations supplied for a particular task.
In diffusion and flow matching models~\cite{lipman2022FlowMatchingGenerative}, inference-time guidance incorporates
these observations into the iterative sampling process, allowing adaptation
without changing the trained model weights.
For example, diffusion posterior sampling uses measurement information to
guide generation from a pretrained diffusion
model~\cite{chung2022DiffusionPosteriorSampling}, and related methods enforce
measurements by inpainting or by correcting the predicted clean sample at each
step~\cite{lugmayr2022RePaintInpaintingUsing,wang2022ZeroshotImageRestoration,song2022PseudoinverseGuidedDiffusionModels}.
Pretrained flow matching models solve inverse problems through gradient-based
guidance~\cite{pokle2024TrainingfreeLinearImage}, optimization of the source
sample~\cite{benhamu2024DFlowDifferentiatingFlows}, plug-and-play alternation
of data-fidelity and denoising steps~\cite{martin2025PnPFlowPlugandplayImage},
or projection of the estimated clean sample onto the measurement-consistent
set~\cite{pourya2025FlowerFlowMatchingSolver}. For time series, a single
unconditionally trained diffusion model can be guided at inference for
forecasting, refinement, and synthesis~\cite{kollovieh2023PredictRefineSynthesize}.
These methods target images and generic time series, whereas monthly smart
meter profiles vary in length and calendar alignment, and DSO observations
range from energy totals and peaks to missing intervals and coarse
readings.

Following this principle, we present \ours, a conditional flow matching model
for monthly smart meter profiles.
Customer and calendar attributes condition the learned profile distribution.
At inference, task-specific observation operators describe monthly statistics,
measured intervals, or coarse readings. Projection-based guidance uses these
operators to steer generated profiles toward the corresponding constraints.
The same trained model weights serve conditional generation, constrained
generation, imputation, and super-resolution.
\ours adapts conditional flow matching and
endpoint projection to these requirements, combining learned customer and
calendar conditioning with task-specific observation operators in one model.

Case studies on Dutch large-customer consumption and photovoltaic generation
profiles and on Belgian residential profiles demonstrate the model across
these tasks.
Without task-specific retraining, \ours achieves lower mean Continuous Ranked
Probability Score (CRPS) than the compared task-specific baselines in all four
evaluated customer categories for imputation and 16-fold super-resolution
(Table~\ref{tab:crps_imputation}).
On a public dataset of 2,400 residential meters, \ours matches the imputation CRPS of the task-trained diffusion model \sssd~\cite{alcaraz2022DiffusionbasedTimeSeries} with an ancestral sampler, lowers its super-resolution CRPS by 4.5\% with the default sampler and more realistic short-term variability, and is trained once in a quarter of the time \sssd needs for one task (Table~\ref{tab:public}).
These results demonstrate that adaptation across observation settings and
high-quality monthly profile modeling can be delivered by one trained model.

The paper makes two contributions:
\begin{itemize}
    \item \textit{Adaptation without task-specific retraining.}
    We formulate constrained generation, imputation, and super-resolution as
    observation-guided sampling tasks for one conditional generative model.
    Changing the observation operator and its associated projection adapts
    the model to the task while keeping the trained weights fixed.

    \item \textit{High-fidelity monthly profile modeling.}
    We adapt the Transformer backbone of \energydiff to conditional flow
    matching for monthly \SI{15}{\min} consumption and photovoltaic generation
    profiles~\cite{lin2025EnergyDiffUniversalTimeseriesa}.
    Calendar-aligned positional encoding and valid-step masking account for
    weekday alignment and variable month lengths.
    We demonstrate fidelity in conditional generation and after observation-guided
    adaptation on large-customer and residential data through distributional
    similarity, probabilistic accuracy, and errors in peak, ramp, and aggregate
    values.
\end{itemize}

\section{Observation-Guided Flow Matching}
\label{sec:flow-matching}
\ours separates the learned monthly profile distribution from the observations
available for a particular task. Customer and calendar attributes condition the
velocity network during training. At inference, observations correct a predicted
complete profile and thereby steer the sampling trajectory. This section first
describes how one task-agnostic velocity model is trained on customer and
calendar attributes, then how observations guide its sampling, and finally the
task-specific corrections.

\subsection{Profiles, Conditions, and Observations}
A monthly profile contains $T$ valid meter intervals, indexed by $k=1,\ldots,T$. If $p_k$ is the
interval-average net power in kW and $\Delta h$ is the interval duration in
hours, its net energy is $E=\Delta h\sum_{k=1}^{T}p_k$ in kWh.
At \SI{15}{\min} resolution, $\Delta h=0.25$ and $T$ ranges from $2688$ to
$2976$ for months of $28$ to $31$ days. For batching, shorter months are padded
to $2976$ intervals. Padded entries carry no data; they are excluded from
training and from all observations (Section~\ref{sec:velocity-network}).
Hereafter,
$x\in\mathbb{R}^{T}$ denotes the profile normalized by a fixed reference power,
$x_k=p_k/p_{\mathrm{ref}}$; observations are expressed in the same coordinates. Positive net power represents consumption and negative
net power represents generation.

Let $c$ collect customer and calendar attributes, including \textit{customer category},
\textit{generation equipment}, \textit{year}, \textit{month}, \textit{month length}, and \textit{first weekday}. The model
learns the conditional profile distribution $q(x\mid c)$ from pairs $(x,c)$.
Additional observations $y$ are described by an operator $f$ acting on the
complete profile:
\begin{equation}
    y=f(x),\qquad
    \Omega(y)=\{z\in\mathbb{R}^{T}:f(z)=y\}.
    \label{eq:observation-set}
\end{equation}
For constrained generation, $f$ extracts specified profile statistics. For
imputation, it selects measured intervals. For super-resolution, it averages
fine-resolution power readings within each measured coarse interval. These
operators change with the task; the learned network weights remain fixed.

\subsection{Learning the Conditional Profile Distribution}
Flow matching learns a velocity field that transports Gaussian noise to the
profile distribution~\cite{lipman2022FlowMatchingGenerative}. Draw
$x_0\sim\mathcal{N}(0,I)$ independently of the training pair $(x_1,c)$, and
flow time $t\sim\mathcal{U}(0,1)$. The linear probability path and its target velocity are
\begin{equation}
    x_t=(1-t)x_0+tx_1,\qquad \dot{x}_t=x_1-x_0.
    \label{eq:cond-path}
\end{equation}
A neural network $u_\theta(x_t,t,c)$ with parameters $\theta$, the velocity
network of Section~\ref{sec:velocity-network}, approximates this target and
is trained by conditional velocity regression:
\begin{equation}
    \mathcal{L}(\theta)=
    \mathbb{E}_{t,x_0,(x_1,c)}
    \left[\left\lVert u_\theta(x_t,t,c)-(x_1-x_0)\right\rVert_2^2\right].
    \label{eq:loss-mse-conditional}
\end{equation}
Padded intervals are excluded from this loss;
Section~\ref{sec:velocity-network} gives its batch form.

The population minimizer of squared-error regression is the conditional mean
of the target velocity. Let
\begin{equation}
    m_t(x_t,c)=\mathbb{E}[X_1\mid X_t=x_t,c]
    \label{eq:endpoint-mean}
\end{equation}
denote the expected complete profile given the current noisy state and the
attributes: the endpoint toward which the flow is heading, averaged over all
profiles consistent with $x_t$. Eliminating $x_0$ from~\eqref{eq:cond-path}
expresses the optimal velocity through this endpoint for $t<1$:
\begin{equation}
    u_t^*(x_t\mid c)=\frac{m_t(x_t,c)-x_t}{1-t}.
    \label{eq:conditional-mean-flow}
\end{equation}
Predicting the velocity is therefore equivalent to predicting the expected
complete profile. Because observations carry information about the complete
profile, observation guidance acts on this endpoint.
After training, categorical conditional generation integrates
\begin{equation}
    \frac{\Dd x_t}{\Dd t}=u_\theta(x_t,t,c),\qquad
    x_0\sim\mathcal{N}(0,I),
    \label{eq:ode}
\end{equation}
from $t=0$ to $t=1$ to obtain a synthetic profile.

\subsection{Observation-Guided Sampling}
\label{sec:inference-time-guidance}
The learned condition $c$ specifies the profile population. The observation
$y$ supplies information about an individual profile at sampling time. Define
$m_t(x_t,c,y)=\mathbb{E}[X_1\mid X_t=x_t,c,y]$. Applying the same regression
identity to the observation-conditioned distribution yields
\begin{equation}
    u_t^*(x_t\mid c,y)=u_t^*(x_t\mid c)
    +\frac{m_t(x_t,c,y)-m_t(x_t,c)}{1-t}.
    \label{eq:guided-velocity}
\end{equation}
Both conditional means retain $c$. Equation~\eqref{eq:guided-velocity}
identifies what exact observation-guided sampling requires: the expected
complete profile given the observations, $m_t(x_t,c,y)$. This quantity is
unavailable in practice: it has no closed form, and computing it from the
trained model would itself require sampling from the observation-conditioned
distribution, which is the problem we are solving. Exact guidance is therefore
infeasible; we approximate $m_t(x_t,c,y)$ at each step by correcting the
network's endpoint estimate with the observations.

At each sampling step, compute
\begin{equation}
    \hat{x}_1=x_t+(1-t)u_\theta(x_t,t,c).
    \label{eq:unbiased-target-est}
\end{equation}
Equation~\eqref{eq:conditional-mean-flow} shows that $\hat{x}_1$ estimates
$m_t(x_t,c)$. A task-specific correction $\Gamma_y$ then produces a revised
endpoint $\tilde{x}_1=\Gamma_y(\hat{x}_1)$, which approximates $m_t(x_t,c,y)$
and is converted back to a velocity:
\begin{equation}
    \tilde{u}_\theta(x_t,t,c,y)=\frac{\Gamma_y(\hat{x}_1)-x_t}{1-t}.
    \label{eq:final-proj-guidance}
\end{equation}
Relative to the unguided velocity, the added term is
$[\Gamma_y(\hat{x}_1)-\hat{x}_1]/(1-t)$: an observation residual in endpoint
space, scaled by the remaining flow time. This endpoint correction follows
inverse-problem methods for diffusion models that refine the estimated clean
sample with the measurements at every
step~\cite{wang2022ZeroshotImageRestoration,song2022PseudoinverseGuidedDiffusionModels}.
\ours adapts it to the flow matching velocity and to smart meter observation
operators.

Section~\ref{sec:task-corrections} specifies $\Gamma_y$ for each task and,
for linear observations, shows that the correction never moves the endpoint
away from the true profile and characterizes exactly what it leaves to the
network.

The velocity network is trained on the linear path~\eqref{eq:cond-path};
observation-guided sampling integrates the corrected flow in the time
$\tau\in[0,1]$ of the variance-preserving (VP)
schedule~\cite{song2020ScorebasedGenerativeModeling}, $x_\tau=\alpha_\tau x_1
+\sigma_\tau x_0$ with $\alpha_\tau=e^{-\beta_\tau/2}$,
$\sigma_\tau=(1-\alpha_\tau^2)^{1/2}$,
$\beta_\tau=\tfrac{1}{2}(1-\tau)^2(\beta_{\max}-\beta_{\min})
+(1-\tau)\beta_{\min}$, $\beta_{\min}=0.1$, and $\beta_{\max}=20$. A scale-time
transformation~\cite{shaul2023BespokeSolversGenerative} transfers the trained
field without retraining: VP time $\tau$ corresponds to the linear-path time
$t_\tau=\alpha_\tau/(\alpha_\tau+\sigma_\tau)$ of equal signal-to-noise ratio,
with $x_\tau=(\alpha_\tau+\sigma_\tau)\,x_{t_\tau}$. The endpoint
estimate~\eqref{eq:unbiased-target-est} is unchanged, and the corrected
endpoint defines the VP velocity
\begin{equation}
    \tilde{u}_\tau=\frac{\dot{\sigma}_\tau}{\sigma_\tau}\,x_\tau
    +\frac{\dot{\alpha}_\tau\sigma_\tau-\alpha_\tau\dot{\sigma}_\tau}
    {\sigma_\tau}\,\Gamma_y(\hat{x}_1),
    \label{eq:vp-guided-velocity}
\end{equation}
which reduces to~\eqref{eq:final-proj-guidance} for $\alpha_\tau=\tau$ and
$\sigma_\tau=1-\tau$.

Algorithm~\ref{alg:projection-based-posterior-sampling} summarizes the
sampler. Each step requires one network evaluation and no differentiation
through the network; the correction is a closed-form operation on the valid
meter intervals, and padding is excluded from the observation operator. At
the final step, the sampler returns the corrected endpoint
$\Gamma_y(\hat{x}_1)$ without an additional network evaluation. Imputed and super-resolved profiles therefore
match their observations exactly.
Figures~\ref{fig:demo-proj-editing} and~\ref{fig:ours-diagram} illustrate the
endpoint correction and the sampling process.

\begin{algorithm}[t]
\caption{Sampling with Observation Guidance}
\label{alg:projection-based-posterior-sampling}
\begin{algorithmic}[1]
    \REQUIRE Trained $u_\theta$, attributes $c$, observations $y$, correction $\Gamma_y$, steps $S$, VP schedule $(\alpha_\tau,\sigma_\tau)$
    \STATE Draw $x\sim\mathcal{N}(0,I)$
    \FOR{$i=0,\ldots,S-1$}
        \STATE $\tau\gets i/S$, $\;t\gets\alpha_\tau/(\alpha_\tau+\sigma_\tau)$, $\;z\gets x/(\alpha_\tau+\sigma_\tau)$
        \STATE $\hat{x}_1\gets z+(1-t)\,u_\theta(z,t,c)$
        \STATE $\tilde{x}_1\gets\Gamma_y(\hat{x}_1)$
        \STATE $x\gets x+\tilde{u}_\tau/S$ with $\tilde{u}_\tau$ from~\eqref{eq:vp-guided-velocity}
    \ENDFOR
    \STATE Return $\tilde{x}_1$ of the final step on its valid intervals
\end{algorithmic}
\end{algorithm}

\begin{figure}[t]
    \centering
    \input{figures/endpoint_correction.tex}
    \caption{Endpoint correction in one sampling step (schematic for imputation). Each profile is drawn as a point: the horizontal axis stands for the missing intervals and the vertical axis for the measured intervals, so the profiles consistent with the observations form the line $\Omega(y)$ (red), on which $p(x\mid c,y)$ is the target distribution. From the current state $x_t$, the velocity network predicts the complete profile $\hat{x}_1$ (orange, dashed); early in sampling this estimate is a smooth average that lies off the data distribution $q(x\mid c)$ (blue). The correction $\Gamma_y$ overwrites the measured intervals with $y$, projecting $\hat{x}_1$ onto $\Omega(y)$, and the Euler step to $x_{t+\Delta t}$ follows the corrected direction toward $\tilde{x}_1$ (black). Grey lines show unguided flow from noise to data. The sampler returns the corrected endpoint of the last step. Geometry from a two-dimensional toy model.}
    \label{fig:demo-proj-editing}
\end{figure}

\begin{figure}[t]
    \centering
    \input{figures/sampling_overview.tex}
    \caption{Observation-guided sampling. At each step, the velocity network, conditioned on the attributes $c$, predicts the complete profile $\hat{x}_1$; the task-specific correction $\Gamma_y$ makes it consistent with the observations $y$ (for imputation, it overwrites the measured intervals); and an Euler step moves the state toward the corrected profile $\tilde{x}_1$, which is returned at the last step. The network is trained once, and only $\Gamma_y$ changes between tasks. The step is drawn in linear-path time $t$; Algorithm~\ref{alg:projection-based-posterior-sampling} carries it out on a VP time grid through the scale-time transformation.}
    \label{fig:ours-diagram}
\end{figure}

\subsection{Task-Specific Observations and Corrections}
\label{sec:task-corrections}
For a full-row-rank linear operator $A$ and noiseless observations $y=Ax$,
the Euclidean projection solves $\min_z\lVert z-v\rVert_2^2$ subject to $Az=y$:
\begin{equation}
    P_{A,y}(v)=v+A^{\mathrm{T}}(AA^{\mathrm{T}})^{-1}(y-Av).
    \label{eq:linear-observation-projection}
\end{equation}
This correction changes the measured component of $v$ while retaining its
component in the null space of $A$. For linear observations,
$\Gamma_y=P_{A,y}$.

Because the true profile $x_1$ lies in $\Omega(y)$, every sample satisfies
\begin{equation}
    \lVert P_{A,y}(\hat{x}_1)-x_1\rVert_2^2
    =\lVert\hat{x}_1-x_1\rVert_2^2
    -\lVert\hat{x}_1-P_{A,y}(\hat{x}_1)\rVert_2^2,
    \label{eq:projection-pythagoras}
\end{equation}
so the corrected endpoint is never farther from the true profile than the
network estimate.

The projection changes only the measured component of the endpoint. The
following decomposition characterizes what exact guidance adds beyond it.

\begin{proposition}
\label{prop:guidance-decomposition}
Let $\Pi=A^{\mathrm{T}}(AA^{\mathrm{T}})^{-1}A$ and $Q=I-\Pi$ be the
orthogonal projections onto the measured component (the row space of $A$) and
the unmeasured component (the null space of $A$). For noiseless linear
observations $y=Ax_1$ and every $x_t$,
\begin{equation}
\begin{split}
    m_t(x_t,c,y)&=P_{A,y}\bigl(m_t(x_t,c)\bigr)+\Delta_t,\\
    \Delta_t&=Q\bigl(m_t(x_t,c,y)-m_t(x_t,c)\bigr).
\end{split}
    \label{eq:guidance-decomposition}
\end{equation}
If $X_1$ given $c$ is Gaussian, then
$\Delta_t=Q\Sigma_tA^{\mathrm{T}}(A\Sigma_tA^{\mathrm{T}})^{-1}
\bigl(y-Am_t(x_t,c)\bigr)$ with $\Sigma_t=\mathrm{Cov}(X_1\mid x_t,c)$.
If $\Pi X_1$ and $QX_1$ are independent given $c$, then $\Delta_t=0$ for all
$t$, and observation-guided sampling with the exact velocity field and exact
integration draws from $p(x_1\mid c,y)$.
\end{proposition}

\begin{IEEEproof}
The observations fix $\Pi X_1=A^{\mathrm{T}}(AA^{\mathrm{T}})^{-1}y$, which is
also the measured component of $P_{A,y}(v)$ for any $v$, while $P_{A,y}$
leaves $Qv$ unchanged; this gives~\eqref{eq:guidance-decomposition}. The
Gaussian form is the conditioning formula for $y=AX_1$. Under independence,
the isotropic source noise keeps $(\Pi X_1,\Pi X_t)$ independent of
$(QX_1,QX_t)$, so the conditional mean of $QX_1$ given $X_t$ does not change
when $y$ is added, and $\Delta_t=0$; the guided velocity~\eqref{eq:final-proj-guidance}
then equals the exact velocity~\eqref{eq:guided-velocity}.
\end{IEEEproof}

Proposition~\ref{prop:guidance-decomposition} separates exact guidance into
two operations: the projection sets the measured component, and the shift
$\Delta_t$ adapts the unmeasured intervals to it. \ours performs the first in
closed form and leaves the second to the network. Without the network, the
projection cannot produce a realistic profile: applied to the source noise, it
returns $\Pi x_1+Qx_0$, which matches the observations exactly but leaves
noise in every unmeasured interval. The profile structure reaches these
intervals through the state. With exact integration, the measured component
of the guided state is $\Pi x_t=t\,\Pi x_1+(1-t)\,\Pi x_0$ whatever the network
predicts, so the state carries the observations, perturbed by noise that
vanishes as $t\to1$. The endpoint estimate passes this information to the
unmeasured intervals: on the linear path,
$Q\,(\partial m_t/\partial x_t)\,\Pi=\tfrac{t}{(1-t)^2}\,Q\Sigma_t\Pi$, the
covariance between unmeasured and measured intervals given the state. A
missing interval therefore moves with the measured intervals it is correlated
with, and the trained network supplies these correlations from the training
profiles.

Exact guidance relies on the same covariance. In the Gaussian case,
$\Delta_t$ applies $Q\Sigma_t$ to the residual of the clean observations,
$y-Am_t(x_t,c)$, whereas the network applies it to their noisy copy in the
state; $\Delta_t$ is the information that the clean observations add beyond
this copy. Although the guided velocity divides by $1-t$, it does not amplify
network errors. With velocity error
$\varepsilon_\theta=u_\theta(x_t,t,c)-u_t^*(x_t\mid c)$, the endpoint estimate
is $\hat{x}_1=m_t(x_t,c)+(1-t)\varepsilon_\theta$, so the guided
velocity~\eqref{eq:final-proj-guidance} differs from the exact
velocity~\eqref{eq:guided-velocity} by
\begin{equation}
    \tilde{u}_\theta-u_t^*(x_t\mid c,y)=Q\varepsilon_\theta-\frac{\Delta_t}{1-t}.
    \label{eq:guided-velocity-error}
\end{equation}
The projection removes the measured part of the network error and passes the
rest through unchanged.

\subsubsection{Imputation}
Let $O=\{\omega_1,\ldots,\omega_{n_O}\}\subseteq\{1,\ldots,T\}$ contain the
observed intervals, and let $I_O$ contain the corresponding rows of the identity
matrix. The observations are
\begin{equation}
    y=f(x)=I_Ox.
    \label{eq:cond-imp}
\end{equation}
The projection replaces the observed values and retains the predicted missing
values:
\begin{equation}
    [\Gamma_y(v)]_k=
    \begin{cases}
        y_j,& k=\omega_j,\\
        v_k,& k\notin O.
    \end{cases}
    \label{eq:imputation-projection}
\end{equation}
The observed index set can change between profiles without retraining the model.

\subsubsection{Super-Resolution}
A coarse reading averages $R$ consecutive fine-resolution power intervals.
Every month length $T$ is a multiple of the $96$ daily intervals, so any factor
$R$ dividing $96$ also divides $T$. Define
$\mathcal{K}_j=\{(j-1)R+1,\ldots,jR\}$ and $D_{jk}=1/R$ for
$k\in\mathcal{K}_j$, with zero entries elsewhere. Then
\begin{equation}
    y=f(x)=Dx.
    \label{eq:cond-sr}
\end{equation}
Writing $\bar{v}_j=R^{-1}\sum_{k\in\mathcal{K}_j}v_k$, the projection is
\begin{equation}
    [\Gamma_y(v)]_k=v_k+y_j-\bar{v}_j,\qquad k\in\mathcal{K}_j.
    \label{eq:super-resolution-projection}
\end{equation}
Each block receives the same offset, preserving its predicted within-block
variations while making its average equal to the measured reading. The factor
$R$ determines the observation operator and leaves the model weights fixed.

\subsubsection{Constrained Generation}
Monthly extrema and mean power are described by
\begin{equation}
    y=f(x)=
    \begin{bmatrix}
        \max_k x_k & T^{-1}\sum_{k=1}^{T}x_k & \min_k x_k
    \end{bmatrix}^{\mathrm{T}}.
    \label{eq:cond-res-gen}
\end{equation}
Denote the corresponding targets by $y_{\max}$, $y_{\mathrm{avg}}$, and
$y_{\min}$. A physical energy total $E$ specifies the mean power $E/(T\Delta h)$,
so $y_{\mathrm{avg}}=E/(T\Delta h\,p_{\mathrm{ref}})$. Extrema are signed values:
$y_{\max}$ is the peak consumption, and $y_{\min}$ is the base load of a
consumption-only customer or the peak generation of a customer with PV.

For the mean constraint, use
$C_{\mathrm{avg}}(v)=v+(y_{\mathrm{avg}}-\bar{v})\boldsymbol{1}$, where
$\bar{v}=T^{-1}\sum_k v_k$. For the maximum constraint, clip values above
$y_{\max}$, then set the entries numerically tied with the largest clipped value
to $y_{\max}$. Denote this map by $C_{\max}$. The minimum correction is
$C_{\min}(v)=-C'(-v)$, where $C'$ is the same maximum-correction rule with
target $-y_{\min}$. Each map is a Euclidean projection onto its own constraint
set, up to tie-breaking for the extrema. Several constraints are combined by
composing their projections; for the joint statistics, the correction applies
the three maps in sequence:
\begin{equation}
    \Gamma_y(v)=C_{\min}\bigl(C_{\mathrm{avg}}(C_{\max}(v))\bigr).
    \label{eq:statistic-correction}
\end{equation}
Unlike the linear case, the conditional mean $m_t(x_t,c,y)$ need not satisfy
extremum constraints: because the maximum is convex, the maximum of the
conditional mean is at most $y_{\max}$. Here $\Gamma_y$ therefore acts as a
feasibility correction rather than an estimate of that mean.

The composition is applied at every flow step. One pass satisfies the minimum
exactly, and the residuals it leaves in the maximum and mean are corrected
again at the following steps. The mean constraint is affine, and once the
extremum locations stabilize, the extremum constraints coincide locally with
polyhedra; each pass then contracts the distance of the endpoint to their
intersection, which lies in $\Omega(y)$, by a factor
$\rho<1$~\cite{hoffman1952ApproximateSolutionsSystems,bauschke1996ProjectionAlgorithmsSolving}.
If the network estimate drifts by at most $\varepsilon$ between steps, the
endpoint approaches $\Omega(y)$ geometrically, up to a distance of
$\rho\varepsilon/(1-\rho)$.

\section{Monthly Profile Velocity Network}
\label{sec:velocity-network}
The velocity network adapts the Transformer backbone of
\energydiff~\cite{lin2025EnergyDiffUniversalTimeseriesa} to conditional flow
matching. Consecutive meter intervals are folded into tokens, embedded in the
model dimension, processed by Transformer blocks, and projected back to profile
coordinates. Flow time and customer attributes condition the blocks. Calendar
alignment and valid-step masking accommodate the monthly observation grid.

\subsection{Calendar-Aligned Positional Encoding}
The same meter index falls on different weekdays across months. Let $w\in
\{0,\ldots,6\}$ denote the first weekday, $N_{\mathrm{day}}=96$ the number of
meter intervals per day, and $r$ the folding factor. For token index
$j=0,\ldots,J-1$, use the shifted position
\begin{equation}
    s_j=j+\left\lfloor\frac{wN_{\mathrm{day}}}{r}\right\rfloor,\qquad
    e_j^{\mathrm{pos}}=\mathrm{MLP}(\mathrm{PE}(s_j)).
    \label{eq:calendar-aligned-encoding}
\end{equation}
Here $\mathrm{PE}$ is a sinusoidal positional encoding~\cite{vaswani2017AttentionAllYou}
and the MLP maps it to the model dimension. Adding $e_j^{\mathrm{pos}}$ to each
token embedding aligns the positional coordinates with the weekly cycle.
The meter index $k$, token index $j$, and flow time $t$ have distinct roles.

\subsection{Valid-Step Masking}
For batching, profiles are padded to $T_{\max}=2976$ intervals. Define the
interval mask $\nu_k=1$ for $k\leq T$ and $\nu_k=0$ otherwise. A folded token is
valid if it contains at least one valid interval. Each attention head adds a
bias of $-\infty$ to the scores of padded profile keys and zero to those of
valid profile and condition-token keys, so valid tokens never attend to
padding. Padding queries retain a self-key to avoid fully masked rows; their
output intervals are discarded.

For a batch $\mathcal{B}$ of padded profiles, the training loss is normalized
by the number of valid meter intervals:
\begin{equation}
    \mathcal{L}_{\mathcal{B}}=
    \frac{\displaystyle\sum_{n\in\mathcal{B}}\sum_{k=1}^{T_{\max}}
    \nu_k^{(n)}\left(u_{\theta,k}^{(n)}-x_{1,k}^{(n)}+x_{0,k}^{(n)}\right)^2}
    {\displaystyle\sum_{n\in\mathcal{B}}\sum_{k=1}^{T_{\max}}\nu_k^{(n)}}.
    \label{eq:masked-flow-loss}
\end{equation}
Here $u_\theta^{(n)}=u_\theta(x_{t^{(n)}}^{(n)},t^{(n)},c^{(n)})$.
Padding therefore contributes neither observation constraints nor regression
error. The generated profile is returned on the $T$ valid intervals.

\subsection{Condition Injection}
Customer and calendar attributes enter through adaptive normalization and
multi-modal attention. A learned embedding of $c$ is combined with the flow-time
embedding to modulate each residual sublayer:
\begin{align}
    Z&=(1+\lambda(t,c))\odot\layernorm(H)+\beta(t,c),\notag\\
    H&\gets H+g(t,c)\odot\mathrm{SubLayer}(Z).
    \label{eq:adaptive-conditioning}
\end{align}
The scale $\lambda$, shift $\beta$, and gate $g$ are learned vectors broadcast
over tokens; $\mathrm{SubLayer}$ denotes attention or the feedforward network.
Multi-modal attention additionally represents the attributes as condition
tokens and computes attention jointly with the profile tokens, enabling
information exchange between the two sequences. These learned mechanisms
specify the population $q(x\mid c)$. Numerical observations $y$ enter through
the sampling correction in Section~\ref{sec:inference-time-guidance}.

\section{Case Study}
\label{sec:case-study}
We evaluate \ours on a DSO dataset of large-customer smart meter profiles and on a public smart meter dataset. This section details the datasets, tasks, metrics, baselines, and model setup.

\subsection{Datasets}
\label{sec:datasets}

\subsubsection{DSO Dataset}
We use a real dataset of monthly smart meter segments from 2022 to 2024, collected by a Dutch DSO from large customers with connections of \SI{60}{\kilo\watt} to \SI{160}{\kilo\watt}. Net power is normalized by a fixed reference power $p_{\mathrm{ref}}=\SI{200}{\kilo\watt}$, which maps the data to $[-1, 1]$ and keeps zero power at zero; positive values indicate consumption and negative values indicate generation. Each customer is assigned one of 26 profile categories, defined by business sector or by the profile operating time (PBT), i.e., the standard annual consumption divided by the contracted transport capacity~\cite{pve2005ProfielenmethodiekElektriciteit}. The model is trained on all categories and evaluated on four of them. E3A, E3B, and E3C group customers with a PBT of at most \num{2000}, \numrange{2000}{3000}, and \numrange{3000}{5000} hours, respectively; the evaluation uses their customers without generation. PV is the category for the consumption profiles of customers with photovoltaic generation; the evaluation uses its customers with large-scale solar generation.

We evaluate on January and July of each year as representative winter and summer months. The dataset is split by customer into training, validation, and test sets with a 70/15/15 ratio, ensuring no customer appears in multiple splits. The training set contains approximately 350,000 monthly segments. The validation set is used solely for overfitting detection.

For the profile-level tasks, the evaluation profiles are representatives of the test set. For each category and evaluated month, the test profiles are partitioned into 100 clusters by k-means on the profile values, and the member with the highest silhouette score represents each cluster. The resulting 100 representative profiles per category and month cover the dominant profile shapes of the test set.

\subsubsection{Public Dataset}
The Fluvius open dataset~\cite{fluvius2025ConsumptionProfilesDigital} contains \SI{15}{\min} offtake and injection energy of 2,400 anonymized residential smart meters in Flanders, Belgium, for 2024. Its eight groups of 300 meters combine the presence of PV generation, a heat pump, and home electric vehicle charging. These small low-voltage loads complement the large customers of the DSO dataset. We form 28,800 monthly net power segments as for the DSO dataset and normalize them by $p_{\mathrm{ref}}=\SI{19}{\kilo\watt}$, the 0.9999 quantile of absolute power in the training split, rounded up; the 0.007\% of training values beyond $\pm p_{\mathrm{ref}}$ are clipped. The meters are split by customer into training, validation, and test sets of 1,680, 360, and 360 meters (70/15/15), stratified by group. The group and the PV indicator take the roles of the customer category and the generation type. The evaluation uses all 360 test customers in January and July, i.e., 720 profiles without selection.

\subsection{Evaluation Tasks}
\label{sec:evaluation-tasks}
The four tasks below are defined on the DSO dataset. In Tasks~2--4, the learned baselines are trained for the task, whereas \ours uses the Task~1 model with the corrections of Section~\ref{sec:task-corrections}. On the public dataset, all four tasks use the 720 evaluation profiles: Task~1 generates ten profiles for the conditions of each profile and tests each group and month, Task~2 uses the maximum, mean, and minimum of each profile with ten samples, and Tasks~3 and~4 draw 100 samples per profile.

\subsubsection{Task 1: Conditional Generation}
\label{sec:task-1-desc}
We generate synthetic data conditioned on year, month, month length, first weekday, customer category, and generation equipment type. As these are abstract conditions, we use only the trained conditional velocity model without inference-time guidance. For evaluation, we generate 1,500 samples per year, month, and customer category combination, totaling 36,000 synthetic profiles. Fidelity is assessed using the Maximum Mean Discrepancy (MMD) permutation test.

\subsubsection{Task 2: Constrained Generation}
In addition to the Task 1 attributes, we specify the monthly maximum, mean, and minimum power of a test profile; the mean determines the monthly net energy. The peak is the maximum for the consumption-only categories and the minimum, i.e., the peak generation, for PV. As these conditions are specific to individual profiles, real samples with identical conditions are too few for an MMD test. We therefore select 99 test profiles as representatives of k-means clusters on their maximum, mean, and minimum, generate 32 samples for each, 3,168 in total, and evaluate the CRPS and the errors of the specified values.

\subsubsection{Task 3: Imputation}
We evaluate missing value imputation under a Missing Not at Random setting with consecutive missing blocks. Block lengths range from 5 to 595 steps, with a nominal missing rate of 0.2; because randomly generated blocks may overlap, the realized rate is at most 0.2 (0.12 to 0.20 on the public dataset).

For each of the four categories and six evaluated months (January and July of 2022--2024), we use the 100 representative test profiles and draw 100 samples for each.

\subsubsection{Task 4: Super-Resolution}
We downsample the representative test profiles by averaging 16 consecutive steps, converting \SI{15}{\min} to \SI{4}{\hour} resolution, then reconstruct the original \SI{15}{\min} profiles. Performance is evaluated using CRPS and Peak Load Error.

\subsection{Evaluation Metrics}

\subsubsection{MMD-based Permutation Testing}
Maximum Mean Discrepancy~\cite{gretton2012KernelTwoSampleTest} measures distributional similarity in kernel space but lacks interpretable thresholds. We therefore apply a permutation test~\cite{hall2002PermutationTestsEquality} following~\cite{lin2025EnergyDiffUniversalTimeseriesa}.

The test compares the observed MMD between the real and synthetic sets with its distribution over random permutations of the pooled samples, which is its distribution under the null hypothesis that both sets come from the same distribution. A small $p$-value indicates that the observed MMD is unusually large, i.e., a distributional mismatch. A large $p$-value means that the test detects no mismatch at the given sample size; it does not establish that the two distributions are equal, and $p$-values do not rank models by fidelity. We therefore use the $p$-value to decide whether a mismatch is detected at the 0.05 level and the MMD value to compare similarity.

\subsubsection{Continuous Ranked Probability Score}
CRPS generalizes the mean absolute error to probabilistic predictions and reduces to it for a point prediction. For real data $(x_i, y_i)$ and a probabilistic model $\hat{p}(x|y)$:
\begin{equation}
    \mathrm{CRPS}(\hat{p}(x|y_i), x_i) = \int_\mathbb{R} (\hat{F}_i(x)-H(x-x_i))^2 \, dx,
\end{equation}
where $\hat{F}_i$ is the CDF of $\hat{p}(x|y_i)$ and $H(\cdot)$ is the Heaviside step function.

\subsubsection{Peak Errors}
Peak power information is often lost in low-resolution profiles. The peak load error (PLE) evaluates two extreme values; to mitigate outlier effects, it uses the 0.9985 and 0.0015 quantiles instead of naive extrema:
\begin{multline}
    \ple{(\hat{p}(x|y_i), x_i)} = \mathrm{CRPS}(\hat{p}(Q_{0.9985}(x)|y_i), Q_{0.9985}(x_i)) \\
    + \mathrm{CRPS}(\hat{p}(Q_{0.0015}(x)|y_i), Q_{0.0015}(x_i)),
\end{multline}
where $Q_\alpha(x)$ denotes the $\alpha$-quantile of profile $x$, and $\hat{p}(Q_\alpha(x)|y_i)$ is the induced distribution obtained by sampling from $\hat{p}(x|y_i)$.
In imputation, these monthly extremes are usually observed. Therefore, we also define the daily peak error (DPE), the absolute error of each daily maximum, $\lvert\max_{k\in d}\tilde{p}_k-\max_{k\in d}p_k\rvert$ in kW for a sample $\tilde{p}$, averaged over samples and over the days $d$ that contain evaluated intervals, i.e., missing intervals in imputation and all intervals in super-resolution. 

\subsubsection{Ramp Ratio and Coverage}
On the public dataset, two further metrics check whether the generated samples are realistic and accurate. For the coverage, the 100 generated samples of a profile give a central 90\% band, from the 5\% to the 95\% quantile, at each evaluated interval; the coverage is the share of real values that fall inside their band. Calibrated samples attain 0.9, and lower values indicate overconfident samples. The ramp ratio checks the changes between consecutive intervals: it divides the mean absolute \SI{15}{\min} change $\lvert p_{k+1}-p_k\rvert$ of the generated samples by that of the real profile over the evaluated intervals and is averaged geometrically over profiles. A value of 1 indicates realistic short-term variability, values below 1 over-smoothed samples, and values above 1 overly variable samples. A difference between two models is called significant when the 95\% bootstrap interval of their paired difference, resampling customers, excludes zero.

\subsection{Baselines for Comparison}
We compare against models that, like \ours, produce a distribution of profiles for given conditions or observations, so that sample-based metrics such as CRPS and MMD apply. Deterministic temporal models such as Autoformer~\cite{wu2021AutoformerDecompositionTransformers} and TimesNet~\cite{wu2023TimesNetTemporal2DVariation} return a single estimate and are trained for one fixed input-output configuration. They neither provide sample distributions nor serve several observation configurations with one trained model, and are therefore not included.

\subsubsection{Task 1}
We compare against a conditional $\beta$-VAE~\cite{higgins2017BetaVAELearningBasic} with 24M parameters and a conditional WGAN~\cite{arjovsky2017WassersteinGenerativeAdversarial} with 76M parameters. Both models use convolutional layers and adaptive layer normalization for condition injection, similar to our flow matching model. Both are trained until convergence.

\subsubsection{Task 2}
For deep learning baselines, we use the same $\beta$-VAE and WGAN models from Task 1, extended with additional conditions: monthly minimum, maximum, and average values.

As a non-learned baseline, an \textit{algebraic scaler} min-max normalizes the average training profile of the customer category, raises it elementwise to a power $a$, and multiplies it by the target maximum $y_m$. The exponent $a$ is solved so that the profile sum equals the target $y_s$, so both targets are met exactly whenever such an exponent exists.

\subsubsection{Task 3}
We use two baselines. First, a Masked Autoencoder~\cite{he2022MaskedAutoencodersAre} with a Transformer encoder-decoder architecture and approximately 2M parameters, trained with MSE loss to predict missing values directly.

Second, we use LoadPIN~\cite{li2024LoadProfileInpainting}, a GAN-based model designed for load profile imputation. LoadPIN consists of a coarse network, a refine network, and a discriminator. We replace the original coarse network with the Masked Autoencoder Transformer described above. Following~\cite{li2024LoadProfileInpainting}, we pre-train the coarse network, then freeze it and train only the refine network and discriminator adversarially. The coarse and refine networks each have approximately 2M parameters; the discriminator has 15k parameters. We use $\lambda_1=\lambda_2=0.1$ in the loss function and a learning rate of 0.001.

\subsubsection{Task 4}
We reproduce ProfileSR~\cite{song2022ProfileSRGANGANBased}, a GAN-based super-resolution model. We use the same architecture as the original but increase hidden dimensions for improved performance. The generator, discriminator, and polish networks have 507k, 23k, and 76k parameters, respectively. Training follows the two-stage procedure in~\cite{song2022ProfileSRGANGANBased} with a learning rate of 0.0002.

\subsubsection{Public Dataset Baselines}
On the public dataset, the $\beta$-VAE and WGAN of Tasks~1 and~2 are retrained with the same architectures and hyperparameters. For Tasks~3 and~4, we train the diffusion model \sssd~\cite{alcaraz2022DiffusionbasedTimeSeries}, built on structured state space layers for time series imputation, separately for each task with its official code and configuration (48M parameters, 200 diffusion steps). A month enters as 16 channels of 186 steps of \SI{4}{\hour}, the patch layout of \ours. For imputation, \sssd is trained on missing blocks drawn as in Task~3. Super-resolution is not supported natively; the conditioning series is the \SI{4}{\hour} average repeated over its 16 intervals. Linear interpolation is the training-free reference. The Masked Autoencoder, LoadPIN, and ProfileSR are used on the DSO data only.

\subsection{Implementation Details}
The velocity network $u_\theta(x_t,t,c)$ has 12 Transformer layers, model dimension $d=128$, feedforward dimension $d_{\text{ff}}=512$, and 7M parameters in total. We train using the Adam optimizer~\cite{kingma2014AdamMethodStochastic} with weight decay, a learning rate of $10^{-4}$, and 500k iterations, corresponding to approximately 267 epochs. Following~\cite{lin2025EnergyDiffUniversalTimeseriesa}, we maintain an exponential moving average (EMA) of the model weights. All baselines requiring training use the same optimizers.

For sampling, we use the EMA weights with 500 Euler steps. In the DSO results
of Tasks 2--4, the final step returns the integrated state. On the public dataset, \ours is trained for 100k iterations with batch size 192 on one NVIDIA H100 GPU in \SI{2.8}{\hour}, with a validation loss that levels off after about 50k iterations without rising, and draws the 72,000 imputation samples in \SI{1.0}{\hour} on the same GPU. On the same GPU, the retrained VAE and WGAN train in \SI{1.2}{\hour} and \SI{2.2}{\hour}, each \sssd model trains in \SIrange{10.8}{10.9}{\hour}, and \sssd draws the 72,000 samples of a task in \SI{4.7}{\hour}. A diffusion model trained with the same network, data, guidance, and sampler reaches an imputation CRPS within 0.3\% of \ours on the public dataset. Because the linear path is Gaussian, the trained model also defines an equivalent diffusion model, which can be sampled with the reverse-time SDE instead of the probability-flow ODE~\cite{song2020ScorebasedGenerativeModeling}. Its ancestral DDPM discretization~\cite{ho2020DenoisingDiffusionProbabilistic}, also used by \sssd, adds fresh noise at every step; with the same grid and endpoint correction, we denote this variant \ours-A.

\section{Results}
\label{sec:results}

\subsection{Task 1: Conditional Generation}
Table~\ref{tab:mmd_cond_gen} reports the MMD permutation tests of Task~1. For \ours, no mismatch is detected at the 0.05 level in any category and month; the lowest $p$-values are $0.0570$ (E3B, July 2022) and $0.0640$ (E3C, January 2024). For the VAE and GAN, a mismatch is detected in every category, with average $p$-values below $0.0001$ except for the GAN in E3C, and their MMD values are 2 to 28 times those of \ours.

\begin{table}[t]
    \centering
    \caption{MMD permutation tests for Task 1 on the DSO dataset, computed per month: average MMD, average $p$-value, and worst-case (lowest) $p$-value over months.}
    \scalebox{0.75}{
    \setlength{\tabcolsep}{2pt}
    \begin{tabular*}{\linewidth}{@{\extracolsep{\fill}}llcccc}
        \toprule
        & Model & E3A & E3B & E3C & PV \\
        \midrule
        \multirow{3}{*}{MMD$^\downarrow$} & \ours & $0.0035$ & $0.0090$ & $0.0119$ & $0.0042$ \\
        & VAE & $0.0983$ & $0.0939$ & $0.0476$ & $0.0764$ \\
        & GAN & $0.0140$ & $0.0239$ & $0.0241$ & $0.0305$ \\
        \midrule
        \multirow{3}{*}{$p$-value} & \ours & $0.3925$ & $0.1612$ & $0.0945$ & $0.5430$ \\
        & VAE & $<0.0001$ & $<0.0001$ & $<0.0001$ & $<0.0001$ \\
        & GAN & $<0.0001$ & $<0.0001$ & $0.0150$ & $<0.0001$ \\
        \midrule
        \multirow{3}{*}{\shortstack[l]{$p$-value\\worst case}} & \ours & $0.2550$ & $0.0570$ & $0.0640$ & $0.3530$ \\
        & VAE & $<0.0001$ & $<0.0001$ & $<0.0001$ & $<0.0001$ \\
        & GAN & $<0.0001$ & $<0.0001$ & $<0.0001$ & $<0.0001$ \\
        \bottomrule
    \end{tabular*}%
    }
    \label{tab:mmd_cond_gen}
    \vspace{-0.4cm}
\end{table}

\subsection{Task 2: Constrained Generation}
\ours uses the Task~1 model with guidance for the three statistics; the errors are evaluated on the monthly peak, i.e., the maximum for consumption and the minimum for PV, and the monthly mean. Table~\ref{tab:crps_res_gen} shows that \ours attains the lowest average CRPS in all four categories and the lowest worst-case CRPS except in E3B, where the GAN is slightly lower. The near-zero best cases stem from a few trivial profiles, such as all-zero records. The scaler has the highest average CRPS, as it only rescales an average profile. Table~\ref{tab:crps_res_gen} also reports the errors of the conditioned values. The scaler matches them exactly by construction, except for the PV peak, where no such exponent exists and only the mean is scaled. Among the learned models, \ours has the lowest peak and mean errors, at least 30\% below the second best, whereas the VAE and GAN can deviate strongly from the specified values.
On the public dataset, where the corrected endpoint is returned, the specified maximum, mean, and minimum are met within \SI{0.01}{\kilo\watt} on average (Section~\ref{sec:results-public}).

\begin{table}[t]
    \centering
    \caption{Constrained generation on the DSO dataset: CRPS of the generated profiles and RMSE of the conditioned monthly peak and mean.}
    \scalebox{0.75}{
    \begin{tabularx}{\linewidth}{lYYYYY}
        \toprule
        \multirow{2}{*}{Method} & \multicolumn{3}{c}{CRPS$^\downarrow$} & \multicolumn{2}{c}{RMSE$^\downarrow$} \\
        & Average & Best & Worst & Peak & Average \\
        \midrule
        \multicolumn{6}{c}{E3A} \\
        \midrule
        Scaler & $0.0325$ & $0.0000$ & $0.1492$ & $0.0000$ & $0.0000$ \\
        VAE & $0.0275$ & $0.0040$ & $0.1287$ & $0.1084$ & $0.0047$ \\
        GAN & $0.0241$ & $0.0080$ & $0.1128$ & $0.0449$ & $0.0175$ \\
        \ours & $0.0185$ & $0.0001$ & $0.0988$ & $0.0177$ & $0.0010$ \\
        \midrule
        \multicolumn{6}{c}{E3B} \\
        \midrule
        Scaler & $0.0315$ & $0.0000$ & $0.1616$ & $0.0000$ & $0.0000$ \\
        VAE & $0.0242$ & $0.0050$ & $0.1172$ & $0.0676$ & $0.0051$ \\
        GAN & $0.0255$ & $0.0083$ & $0.0993$ & $0.0433$ & $0.0175$ \\
        \ours & $0.0192$ & $0.0002$ & $0.1051$ & $0.0125$ & $0.0011$ \\
        \midrule
        \multicolumn{6}{c}{E3C} \\
        \midrule
        Scaler & $0.0545$ & $0.0136$ & $0.1136$ & $0.0000$ & $0.0000$ \\
        VAE & $0.0291$ & $0.0085$ & $0.0707$ & $0.1036$ & $0.0056$ \\
        GAN & $0.0285$ & $0.0116$ & $0.0582$ & $0.0371$ & $0.0221$ \\
        \ours & $0.0190$ & $0.0005$ & $0.0469$ & $0.0209$ & $0.0015$ \\
        \midrule
        \multicolumn{6}{c}{PV} \\
        \midrule
        Scaler$^*$ & $0.0460$ & $0.0000$ & $0.1403$ & n.a. & $0.0000$ \\
        VAE & $0.0307$ & $0.0048$ & $0.0822$ & $0.0548$ & $0.0086$ \\
        GAN & $0.0283$ & $0.0068$ & $0.0881$ & $0.0362$ & $0.0176$ \\
        \ours & $0.0213$ & $0.0002$ & $0.0736$ & $0.0252$ & $0.0014$ \\
        \bottomrule
        \multicolumn{6}{l}{$^*$: No exponent $a$ exists; only the mean is scaled.}
    \end{tabularx}%
    }
    \label{tab:crps_res_gen}
    \vspace{-0.4cm}
\end{table}

\subsection{Task 3: Imputation}
Table~\ref{tab:crps_imputation} presents the imputation CRPS in each customer category. In every category, \ours lowers the average CRPS of both learned baselines by 41\% to 60\%.

\begin{table}[t]
    \centering
    \caption{Imputation and 16-fold super-resolution on the DSO dataset: average and worst case over the evaluated profiles. Masked AE: masked autoencoder.}
    \scalebox{0.75}{
    \setlength{\tabcolsep}{3pt}
    \begin{tabular*}{\linewidth}{@{\extracolsep{\fill}}lcccccc}
        \toprule
        & \multicolumn{2}{c}{Imputation} & \multicolumn{4}{c}{Super-resolution} \\
        \cmidrule(lr){2-3}\cmidrule(lr){4-7}
        & \multicolumn{2}{c}{CRPS$^\downarrow$} & \multicolumn{2}{c}{CRPS$^\downarrow$} & \multicolumn{2}{c}{PLE$^\downarrow$} \\
        \cmidrule(lr){2-3}\cmidrule(lr){4-5}\cmidrule(lr){6-7}
        Method & Avg. & Worst & Avg. & Worst & Avg. & Worst \\
        \midrule
        \multicolumn{7}{c}{E3A} \\
        \midrule
        Linear Interp. & $0.0623$ & $0.2464$ & $0.0210$ & $0.0662$ & $0.0705$ & $0.4274$ \\
        Masked AE & $0.0344$ & $0.1504$ & -- & -- & -- & -- \\
        LoadPIN & $0.0343$ & $0.2248$ & -- & -- & -- & -- \\
        ProfileSR & -- & -- & $0.0187$ & $0.0656$ & $0.0426$ & $0.3635$ \\
        \ours & $0.0201$ & $0.1757$ & $0.0124$ & $0.0501$ & $0.0321$ & $0.3873$ \\
        \midrule
        \multicolumn{7}{c}{E3B} \\
        \midrule
        Linear Interp. & $0.0594$ & $0.3956$ & $0.0193$ & $0.0993$ & $0.0573$ & $0.4078$ \\
        Masked AE & $0.0333$ & $0.2219$ & -- & -- & -- & -- \\
        LoadPIN & $0.0328$ & $0.1769$ & -- & -- & -- & -- \\
        ProfileSR & -- & -- & $0.0177$ & $0.0774$ & $0.0332$ & $0.3357$ \\
        \ours & $0.0166$ & $0.1098$ & $0.0115$ & $0.0561$ & $0.0255$ & $0.3518$ \\
        \midrule
        \multicolumn{7}{c}{E3C} \\
        \midrule
        Linear Interp. & $0.0584$ & $0.2210$ & $0.0198$ & $0.0654$ & $0.0683$ & $0.2586$ \\
        Masked AE & $0.0339$ & $0.1101$ & -- & -- & -- & -- \\
        LoadPIN & $0.0336$ & $0.1034$ & -- & -- & -- & -- \\
        ProfileSR & -- & -- & $0.0187$ & $0.0657$ & $0.0382$ & $0.1996$ \\
        \ours & $0.0135$ & $0.0525$ & $0.0116$ & $0.0549$ & $0.0248$ & $0.1595$ \\
        \midrule
        \multicolumn{7}{c}{PV} \\
        \midrule
        Linear Interp. & $0.1149$ & $0.4023$ & $0.0378$ & $0.0986$ & $0.1385$ & $0.5090$ \\
        Masked AE & $0.0507$ & $0.1476$ & -- & -- & -- & -- \\
        LoadPIN & $0.0512$ & $0.1587$ & -- & -- & -- & -- \\
        ProfileSR & -- & -- & $0.0334$ & $0.0854$ & $0.0734$ & $0.4154$ \\
        \ours & $0.0285$ & $0.1171$ & $0.0201$ & $0.0608$ & $0.0389$ & $0.3962$ \\
        \bottomrule
    \end{tabular*}%
    }
    \label{tab:crps_imputation}
    \vspace{-0.4cm}
\end{table}

\subsection{Task 4: Super-Resolution}
Table~\ref{tab:crps_imputation} also reports super-resolution. Interpolation is more competitive here than in imputation because only the variation within each coarse interval is missing. \ours achieves the lowest average and worst-case CRPS in each customer category.

Short peaks are lost in coarse readings, so Table~\ref{tab:crps_imputation} also reports the PLE.
Despite its dedicated training to minimize the peak load error, ProfileSR has a higher average PLE than \ours in all four categories, with the largest gap in the PV category; its worst-case PLE is lower in E3A and E3B ($0.3635$ and $0.3357$ versus $0.3873$ and $0.3518$). Without task-specific training, \ours thus achieves superior or comparable performance in super-resolution.

\subsection{Public Dataset}
\label{sec:results-public}
Table~\ref{tab:public} reports the four tasks on the 720 held-out residential profiles, with CRPS and errors in \si{\kilo\watt}. In Task~1, \ours has the lowest MMD, and a mismatch is detected in 1 of the 16 group-month cells, against all 16 for the VAE and WGAN. In Task~2, guidance meets the specified maximum and mean within \SI{0.01}{\kilo\watt} on average, whereas the VAE and WGAN miss the maximum by \SI{3.3}{\kilo\watt} and \SI{2.4}{\kilo\watt}; \ours lowers their CRPS by 30\% and 22\%.

\begin{figure}[t]
    \centering
    \includegraphics[width=\linewidth]{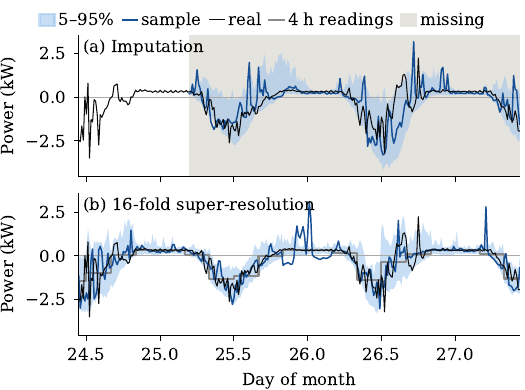}
    \vspace{-0.5cm}
    \caption{Imputation (a) and 16-fold super-resolution (b) of the same held-out household with PV, a heat pump, and home EV charging in July by \ours with fixed weights: real profile, 5--95\% band of 100 samples, one sample, and the 4~h readings in (b); shaded intervals are missing. The CRPS, \SI{0.287}{\kilo\watt} (a) and \SI{0.234}{\kilo\watt} (b), lies at the 54th and 56th percentiles of the 720 profiles.}
    \label{fig:imp_public}
\end{figure}

In imputation, \ours reduces the CRPS of linear interpolation by 60\% and the DPE by 35\%. Its imputed intervals retain 93\% of the real mean \SI{15}{\min} ramp, whereas interpolation retains 2\%. Sampled without observations, i.e., from the customer and calendar labels only with ten samples per profile, the same model has a CRPS of \SI{0.534}{\kilo\watt} on the missing intervals, against \SI{0.334}{\kilo\watt} with guidance.
The task-trained \sssd attains a 3.4\% lower imputation CRPS, a significant difference, and a lower DPE, but its samples are over-smoothed: they retain 71\% of the real mean ramp, against 93\% for \ours. In the example of Fig.~\ref{fig:imp_public}a, the sample of \ours follows the daily PV pattern across the missing days, and the band covers 93\% of the real values. In super-resolution, \ours attains a CRPS 4.5\% below \sssd, also a significant difference, and 36\% below linear interpolation. The PLE of the two learned models does not differ significantly, and \sssd has the lower DPE (\SI{0.85}{\kilo\watt} versus \SI{0.99}{\kilo\watt}); the samples of \ours reproduce the real ramp magnitude (ratio 1.00), whereas those of \sssd retain 64\%. For the same household, the samples reconstructed from 4~h readings (Fig.~\ref{fig:imp_public}b) retain the short-term variability of the real profile (ramp ratio 1.02).
\mbox{\ours-A}, using the same weights and an ancestral sampler like \sssd, match \sssd in imputation, with no significant difference in CRPS, and attain the lowest super-resolution CRPS, \SI{0.264}{\kilo\watt}, with ramp ratios of 0.70 and 0.63, close to those of \sssd, and a coverage of about 0.74 in both tasks. The sampler is thus a choice at inference without retraining: the ancestral sampler gives lower CRPS and, in imputation, a lower DPE, whereas the Euler sampler gives calibrated samples with realistic ramps.

\begin{table}[t]
    \centering
    \caption{Results on the public dataset.}
    \scalebox{0.75}{
    \setlength{\tabcolsep}{2pt}
    \begin{tabularx}{\linewidth}{lYYYY}
        \toprule
        \multicolumn{5}{c}{Task 1: Conditional generation} \\
        \midrule
        Method & MMD$^\downarrow$ & Mismatches$^\downarrow$ & & \\
        \midrule
        \ours & $0.027$ & 1/16 & & \\
        VAE & $0.146$ & 16/16 & & \\
        WGAN & $0.044$ & 16/16 & & \\
        \midrule
        \multicolumn{5}{c}{Task 2: Constrained generation} \\
        \midrule
        Method & CRPS$^\downarrow$ & Max.\ error$^\downarrow$ & Mean error$^\downarrow$ & \\
        \midrule
        \ours & $0.474$ & $0.009$ & $0.004$ & \\
        VAE & $0.676$ & $3.271$ & $0.137$ & \\
        WGAN & $0.606$ & $2.416$ & $0.194$ & \\
        \midrule
        \multicolumn{5}{c}{Task 3: Imputation} \\
        \midrule
        Method & CRPS$^\downarrow$ & DPE$^\downarrow$ & Ramp ratio & Coverage \\
        \midrule
        \ours & $0.334$ & $0.811$ & $0.934$ & $0.848$ \\
        \ours-A & $0.321$ & $0.728$ & $0.701$ & $0.745$ \\
        \sssd & $0.322$ & $0.676$ & $0.710$ & $0.707$ \\
        Linear Interp. & $0.829$ & $1.238$ & $0.017$ & -- \\
        \midrule
        \multicolumn{5}{c}{Task 4: 16-fold super-resolution} \\
        \midrule
        Method & CRPS$^\downarrow$ & PLE$^\downarrow$ & Ramp ratio & Coverage \\
        \midrule
        \ours & $0.270$ & $0.783$ & $1.002$ & $0.889$ \\
        \ours-A & $0.264$ & $0.791$ & $0.635$ & $0.743$ \\
        \sssd & $0.282$ & $0.767$ & $0.643$ & $0.574$ \\
        Linear Interp. & $0.420$ & $2.425$ & $0.117$ & -- \\
        \bottomrule
        \multicolumn{5}{>{\raggedright\arraybackslash}p{0.97\linewidth}}{CRPS, DPE, PLE, and the Task~2 errors in kW. MMD: average over groups and months; Mismatches: group-month cells with $p<0.05$. \ours-A: same weights, ancestral sampler.}
    \end{tabularx}%
    }
    \label{tab:public}
    \vspace{-0.4cm}
\end{table}

\section{Conclusion}
We propose \ours, a novel framework that adapts conditional flow matching and projection-based guidance to monthly smart meter profiles: one model, trained for conditional generation, serves constrained generation, imputation, and super-resolution by changing only the observation operator. On the DSO dataset, it achieves lower mean CRPS than the task-specific baselines for imputation and 16-fold super-resolution in all four categories, and the lowest MMD in conditional generation. On a public residential dataset, it outperforms VAE and WGAN models in conditional and constrained generation, matches the imputation CRPS of a task-trained diffusion model with an ancestral sampler, and improves on its super-resolution CRPS, while its single training takes a quarter of the time that model needs for each task. The sampler is chosen at inference, trading calibrated samples with realistic short-term variability against lower CRPS without retraining. One trained model thus provides realistic profiles across the changing observation settings of DSO practice.

Several directions follow from these results. Projection guidance samples the
observation-conditioned distribution exactly only when measured and unmeasured
intervals are independent (Proposition~\ref{prop:guidance-decomposition});
corrections that use the learned dependence between them could sharpen the
guided distribution where intervals are strongly coupled. On the public
dataset, the task-trained \sssd attains lower daily peak errors than \ours with
either sampler and a lower imputation CRPS than the default Euler sampler, at
the cost of over-smoothed samples and a separate model per task; observation
operators for daily peak statistics, or samplers between the Euler and
ancestral updates, could narrow this gap without task-specific training.
Finally, the observation operators treat measurements as exact; extending them
to noisy or anomalous meter readings would broaden the DSO data the model can
serve.

\bibliographystyle{IEEEtran}
\bibliography{energy_system_planning, energy_transition, machine_learning}

\end{document}